\documentclass[sigconf]{acmart}
\setkeys{acmart.cls}{balance=false}
\usepackage{balance}
\usepackage{amsmath}
\usepackage{tabularx}
\usepackage{multirow}
\usepackage{booktabs}
\usepackage{graphicx}
\usepackage{multirow}
\usepackage{booktabs}
\usepackage{xcolor}
\usepackage{cleveref}

\definecolor{workblue}{RGB}{35,90,160}
\definecolor{workgreen}{RGB}{0,125,75}
\graphicspath{{fig/}{../fig/}}
\setcopyright{none}

\title{AI Meets Formal Methods in the Era of  Vibe Hardware Development}
\title{Agentic AI Meets Formal Verification in the Era of \\ Agentic Hardware Development}
\subtitle{Invited Paper}
\author{Wenji Fang\texorpdfstring{$^\dagger$}{}, Yuchao Wu\texorpdfstring{$^\dagger$}{}, Jing Wang, Zhiyao Xie$^*$}
\affiliation{%
  \institution{Hong Kong University of Science and Technology}
  \city{}
  \country{}
}

\copyrightyear{2027}
\acmYear{2027}
\setcopyright{cc}
\setcctype{by}
\acmConference[ASPDAC '27]{32nd Asia and South Pacific Design Automation Conference}{January 25--28, 2027}{Tokyo, Japan}
\acmBooktitle{32nd Asia and South Pacific Design Automation Conference (ASPDAC '27), January 25--28, 2027, Tokyo, Japan}
\acmDOI{10.1145/3840404.3860940}
\acmISBN{979-8-4007-2923-2/2027/01}

\begin{document}
\begin{abstract}

As AI agents become increasingly capable of generating and optimizing hardware circuits, ensuring the correctness of AI-generated designs becomes more important.
Meanwhile, advances in AI lower the barriers to developing and applying formal methods, creating new opportunities to automate and improve hardware verification.
This invited paper presents an overview of the methodologies and applications of hardware formal methods in the AI era.
First, we examine how agentic AI advances hardware formal methods across three levels: specification, proof construction, and reasoning engines.
Second, we discuss how formal methods support reliable AI-driven hardware development through RTL generation, debugging, and correctness-preserving optimization.
We review representative research alongside our recent contributions, highlighting key technical advances and insights from experimental results.
Finally, we discuss future research opportunities toward proof-native hardware development, where formal reasoning could become an integral part of autonomous hardware design.
\end{abstract}
\maketitle
\pagestyle{empty}

\begingroup\renewcommand\thefootnote{$\dagger$}
\footnotetext{Equal Contribution}
\endgroup

\begingroup\renewcommand\thefootnote{$*$}
\footnotetext{Corresponding Author (eezhiyao@ust.hk)}
\endgroup

\thispagestyle{empty}

\section{Introduction}
\label{sec:introduction}

\begin{figure}[t]
\centering
\includegraphics[width=\linewidth]{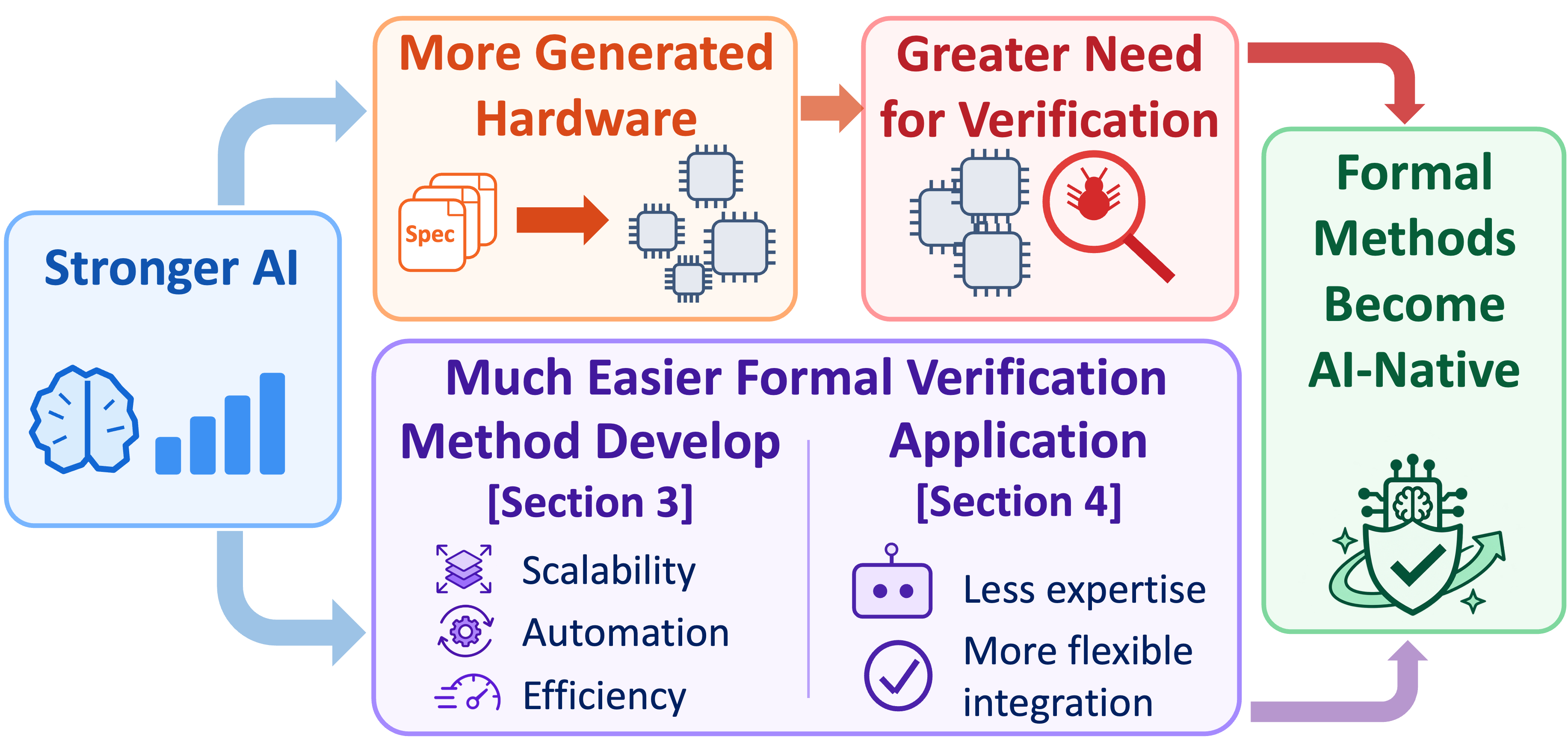}
\vspace{-.2in}
\caption{Advances in AI make hardware generation and optimization more accessible, increasing the demand for formal verification. Meanwhile, AI lowers the barriers to developing and applying formal methods, creating opportunities to integrate formal reasoning into agentic hardware development.}
\label{fig:overview}
\vspace{-.2in}
\end{figure}

\textbf{Recent advances in large language models (LLMs) and AI agents are reshaping hardware development.}
Designers can increasingly express requirements in natural language and rely on AI agents to generate, debug, and optimize RTL implementations.
We refer to this emerging paradigm as \emph{agentic hardware development}.
As AI capabilities improve, agents can explore larger design spaces and perform increasingly complex design tasks with less human intervention.
This trend creates new opportunities for more automated and efficient hardware development.

\textbf{Growing AI capabilities increase the demand for formal verification while creating new opportunities to advance formal methods.}
As shown in~\Cref{fig:overview}, AI agents can generate and modify more hardware designs, increasing the number of design decisions that require correctness checking.
Formal methods provide rigorous guarantees that implementations satisfy their formal specifications.
Meanwhile, developing and applying formal methods traditionally requires substantial expertise in specification, proof construction, and algorithm design.
Advances in AI can reduce these barriers by automating verification tasks, guiding formal reasoning, and accelerating the development of verification algorithms.\looseness=-1

\textbf{This invited paper reviews the methodology and applications of hardware formal methods in the AI era through two perspectives:}
(1) \emph{Part I (Methodology): Agentic AI for Hardware Formal Methods} examines how AI advances formal verification methodologies, improving their automation, efficiency, and scalability.
(2) \emph{Part II (Application): Formal Methods for Agentic Hardware Development} examines how formal verification supports reliable AI-driven hardware generation, debugging, and optimization.
Together, these parts provide a systematic view of recent progress and emerging opportunities for hardware formal methods.

\textbf{Part I (Methodology) organizes recent advances in agentic AI for hardware formal methods into three levels: specification, proof construction, and reasoning engines.}
(1) \emph{Specification:} AI supports assertion generation, specification quality evaluation, and the development of executable behavioral specifications.
(2) \emph{Proof construction:} AI agents guide model checking and automate theorem proving through semantic guidance, lemma generation, and proof reuse.
(3) \emph{Reasoning engines:} AI agents discover and optimize reusable algorithms and heuristics for SAT solving and model checking.
These levels cover the development of correctness specifications, the construction of proofs, and the improvement of underlying reasoning algorithms.

\textbf{Part II (Application) examines three applications of formal methods in agentic hardware development: generation, debugging, and optimization.}
(1) \emph{Generation:} Formal properties, executable models, and proof feedback guide RTL construction and correction.
(2) \emph{Debugging:} Formal counterexamples and verification results support root-cause diagnosis and targeted repair.
(3) \emph{Optimization:} Formal equivalence checking enables correctness-preserving transformations guided by PPA feedback.
These applications demonstrate how formal verification can provide continuous correctness feedback throughout AI-driven hardware development.

\begin{figure}[t]
\centering
\includegraphics[width=1\linewidth]{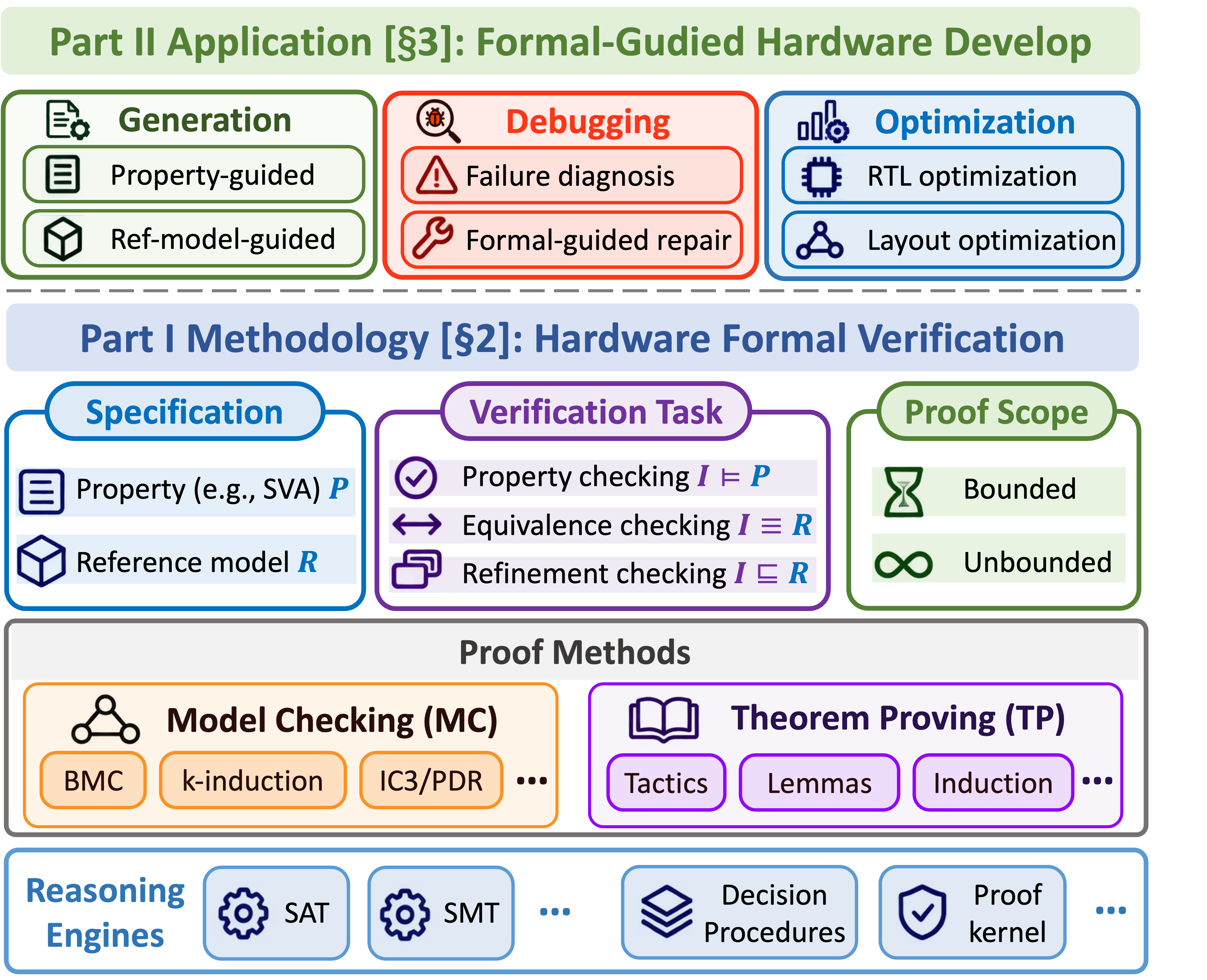}
\vspace{-.25in}
\caption{Overview of our taxonomy of hardware formal verification methodology and application.}
\label{fig:formal_taxonomy}
\label{fig:formal}
\vspace{-.2in}
\end{figure}

\textbf{We present a systematic review of recent advances, supported by representative research, experimental results, and our own contributions.}
Our main contributions are summarized as follows:
\begin{itemize}
    \item \textbf{Methodology: Agentic AI for Hardware Formal Methods.}
    We present a taxonomy of AI-driven advances in formal specification, proof construction, and reasoning engines, highlighting their technical progress and challenges.

    \item \textbf{Application: Formal Methods for Agentic Hardware Development.}
    We organize formal-guided approaches to RTL generation, debugging, and optimization, examining their roles in improving hardware design reliability.

    \item \textbf{Research Progress and Insights.}
    We review representative studies and our recent contributions, using experimental results and case studies to identify key advances and remaining challenges.

    \item \textbf{Future Research Directions.}
    We discuss opportunities in intent-aligned specifications, scalable and reusable proving, and proof-native hardware development.
\end{itemize}

\textbf{We envision formal methods becoming a native part of AI-driven hardware development.}
Advances in requirements engineering can support more complete and intent-aligned formal specifications, while AI-native verification techniques can improve proof automation, efficiency, and scalability.
Integrating these capabilities into hardware generation, debugging, and optimization can enable proof-native development, where AI agents make increasingly autonomous design decisions supported by machine-checkable correctness guarantees.

\section{Part I (Methodology): Agentic AI for Hardware Formal Methods}
\label{sec:ai_for_formal}

\begin{figure*}[!t]
\centering
\vspace{-.2in}
\includegraphics[width=0.9\linewidth]{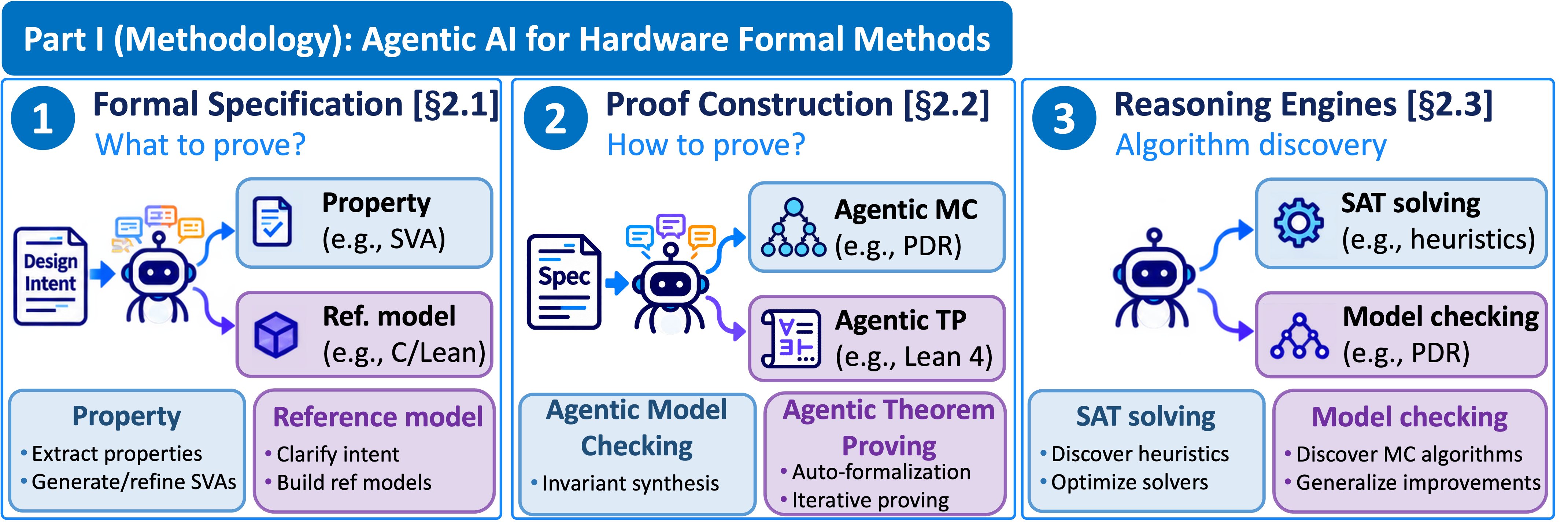}
\vspace{-.15in}
\caption{Three levels of agentic AI for hardware formal methods.
AI supports specification generation and intent clarification, guides task-specific proof construction through model checking and theorem proving, and improves reusable reasoning algorithms through agentic discovery.}
\vspace{-.15in}
\label{fig:part1}
\end{figure*}

Hardware formal verification involves three closely related levels: specifying the intended behavior, constructing proofs against the specification, and executing the underlying reasoning algorithms.
Each level presents a distinct challenge and opportunity for agentic AI.
As illustrated in~\Cref{fig:part1},
(1) \emph{Formal specification} (\Cref{sec:artifacts}) defines what needs to be verified, with AI helping generate and validate correctness targets aligned with designer intent;
(2) \emph{Proof construction} (\Cref{sec:proof}) determines how to establish correctness for a given verification task, with AI guiding model checking and theorem proving through semantic hints, invariants, and lemmas;
and (3) \emph{Reasoning engines} (\Cref{sec:engines}) provide the underlying computational capabilities, with AI discovering reusable algorithms and heuristics to improve solving efficiency.
These three levels connect the correctness target, task-specific proof strategies, and general-purpose reasoning algorithms.
Table~\ref{tab:ai_for_formal} summarizes representative research in these directions.

\subsection{Formal Specification: Capturing Design Intent}
\label{sec:artifacts}

AI-assisted hardware specification is evolving from automated assertion generation toward more complete and reliable formal specifications.
A central challenge is to establish proof targets that accurately capture designer intent and provide sufficient behavioral coverage.
We discuss two complementary directions: property-based and reference model-based specification.

\subsubsection{Property-based Specification}
\label{sec:property_spec}

Property-based specification expresses functional and temporal requirements through formal assertions.
Recent research has progressed from automated assertion generation toward improving the quality, coverage, and bug-detection effectiveness of generated properties.

\begin{table}[t]
\centering
\caption{Our AssertLLM2~\cite{assertllm2_26} benchmark statistics and evaluation metrics.}
\vspace{-.15in}
\label{tab:assertllm2_overview}
\small
\renewcommand{\arraystretch}{1.15}
\setlength{\tabcolsep}{4pt}
\resizebox{\columnwidth}{!}{%
\begin{tabular}{@{}ll|ll@{}}
\toprule
\textbf{Statistic} & \textbf{Value} &
\textbf{Eval. Dimension} & \textbf{Metric} \\
\midrule
Design Scale & 83 & Syntax Check & Compile Pass \\
Func. Categories & 13 & Formal Provability & FPV Proven Rate \\
Avg. RTL Scale & 2.9k LoC & Coverage & COI/Formal/Proof \\
Avg. Spec Scale & 10k tokens & Mutation Testing & Mutant Kill Rate \\
\bottomrule
\end{tabular}%
}
\end{table}

\textbf{Traditional Background.}
Property-based verification expresses functional and temporal requirements through assertions, such as SystemVerilog Assertions (SVA)~\cite{harris2016glast,germiniani2022harm}.
As illustrated in~\Cref{fig:formal_taxonomy}, formal property verification (FPV) checks whether an implementation $I$ satisfies a property $P$, written as $I \models P$.
Traditionally, engineers manually translate design requirements into assertions, requiring expertise in hardware behavior and temporal logic.
The effectiveness of verification depends on the correctness and behavioral coverage of these properties.

\textbf{AI-based Methods.}
Early research explored automating assertion generation from natural-language specifications.
Our AssertLLM framework was among the early efforts to generate SVA from complete specification documents, extracting functional and temporal requirements from text and waveform diagrams~\cite{assertllm25}.
Recent approaches further improve generation through learned grammars, retrieval, specialized language models, and RTL-aware specification analysis~\cite{harris2016glast,hybridnl2sva25,fu2026chatsva,assertionforge25,autosva2_23}.

As assertion generation becomes increasingly automated, evaluating the quality of generated properties becomes a central challenge.
Syntactic correctness and formal provability alone provide limited evidence of verification effectiveness.
Coverage and bug-detection capability offer additional measures of assertion quality.
Coverage-guided approaches such as LASSO~\cite{ankireddy2025lasso} and CoverAssert~\cite{wang2026coverassert} use verification feedback to expand property coverage.
Benchmarks including FVEval~\cite{kang2025fveval} and AssertionBench support systematic evaluation of generated assertions~\cite{pulavarthi2025assertionbench}.

\textbf{Our Work and Insights.}
Our AssertLLM2~\cite{assertllm2_26} benchmark evaluates assertion quality across 83 designs through syntax checking, formal verification, coverage analysis, and mutation testing.
Table~\ref{tab:assertllm2_overview} summarizes its statistics and evaluation metrics.
Claude Sonnet 4.5 achieves 93.04\% syntax success and 65.65\% structural reach, but only 23.30\% proof coverage and an 11.69\% bug-kill ratio.
These results reveal a substantial gap between generating valid assertions and achieving effective verification.
They motivate coverage-driven generation and evaluation methods that measure how well properties detect incorrect hardware behavior.

\subsubsection{Reference Model-based Specification}
\label{sec:reference_spec}

Reference model-based specification describes expected hardware behavior through executable models.
It provides a broader correctness target for equivalence and refinement checking, complementing the selected behaviors expressed by individual assertions.

\begin{figure}[t]
\centering
\includegraphics[width=1\linewidth]{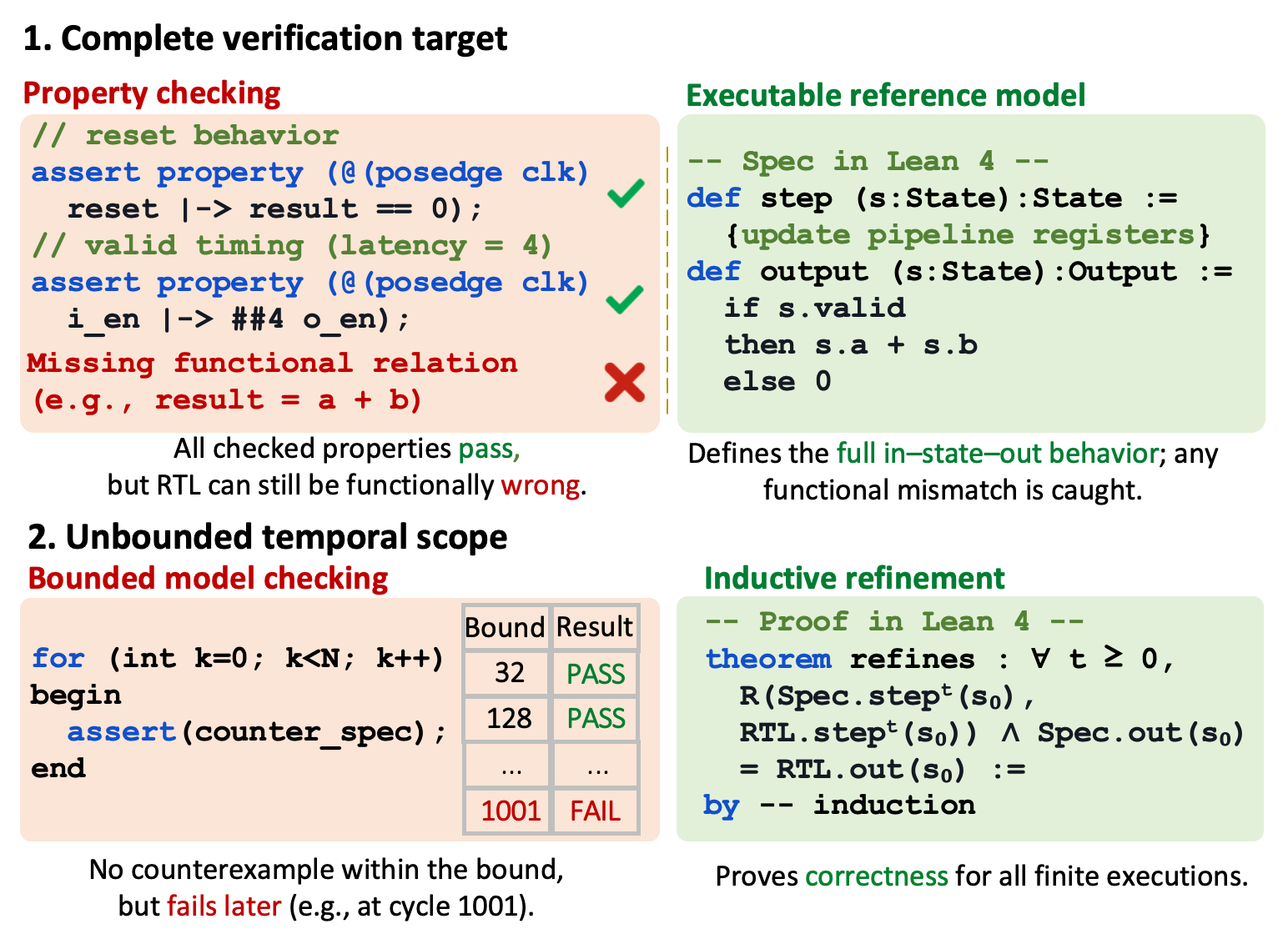}
\vspace{-.3in}
\caption{Two key dimensions of hardware formal verification: specification completeness and proof completeness.
Executable reference models can capture broader intended behavior, while unbounded proofs establish correctness across all execution depths.}
\vspace{-.3in}
\label{fig:spec_proof_comparison}
\end{figure}

\textbf{Traditional Background.}
Reference model-based verification describes expected hardware behavior through executable models, such as RTL, C/C++, or Lean.
As illustrated in~\Cref{fig:formal_taxonomy}, formal equivalence verification (FEV), including combinational equivalence checking (CEC) and sequential equivalence checking (SEC), establishes behavioral equivalence between an implementation $I$ and a reference $R$, written as $I \equiv R$~\cite{van2000sequential,mishchenko2006improvements,li2023se3}.
Refinement checking establishes that an implementation conforms to an abstract reference model, written as $I \sqsubseteq R$~\cite{kami17,fang2023r,xing2022compositional}.
These approaches provide broader behavioral correctness targets, but constructing reliable reference models requires accurately defining state transitions, interfaces, timing, and functional semantics.

\definecolor{workblue}{RGB}{35,90,160}

\begin{table*}[t]
\centering
\vspace{-.2in}
\caption{Representative research in agentic AI for hardware formal methods.}
\label{tab:ai_for_formal}
\vspace{-.1in}

\small
\setlength{\tabcolsep}{4pt}
\renewcommand{\arraystretch}{1.15}

\begin{tabular}{@{}
p{0.11\textwidth}
p{0.12\textwidth}
p{0.21\textwidth}
p{0.51\textwidth}
@{}}
\toprule
\textbf{Category} &
\textbf{Direction} &
\textbf{Representative Work} &
\textbf{Key Contribution} \\
\midrule

\multirow{7}{*}{\textbf{\shortstack[l]{Formal\\Specification}}}
& \multirow{5}{*}{\shortstack[l]{Property-\\based}}
& \textcolor{workblue}{AssertLLM~\cite{assertllm25}}
& Automates SVA generation from specification documents and waveform. \\

& & \textcolor{workblue}{AssertLLM2~\cite{assertllm2_26}}
& Evaluates assertion quality via formal verification, coverage, and mutation testing. \\

& & AssertionForge~\cite{assertionforge25}
& Combines specification and RTL context for assertion generation. \\

& & LASSO~\cite{ankireddy2025lasso}
& Uses formal feedback to improve assertion coverage. \\

& & FVEval~\cite{kang2025fveval}
& Benchmarks assertion generation using syntax and formal verification. \\

\cmidrule(lr){2-4}

& \multirow{2}{*}{\shortstack[l]{Reference\\model-based}}
& FormalRTL~\cite{formalrtl26}
& Uses a supplied C reference model to guide RTL generation and equivalence-based correction. \\

& & HINT~\cite{hint26}
& Introduces an executable hardware-intent representation with specification-level checking and explicit RTL-lowering obligations. \\

\midrule

\multirow{6}{*}{\textbf{\shortstack[l]{Proof\\Construction}}}
& \multirow{3}{*}{Agentic MC}
& LLM4PDR~\cite{llm4pdr26}
& Generates semantic predicates and clauses to guide word-level PDR. \\

& & CIll~\cite{cill26}
& Iteratively generates supporting invariants from counterexamples to induction. \\

& & Agentic-IC3~\cite{agenticic3_26}
& Uses evolving proof states and solver feedback to guide IC3 proof search. \\

\cmidrule(lr){2-4}

& \multirow{3}{*}{Agentic TP}
& CktFormalizer~\cite{cktformalizer26}
& Combines typed circuit representations with interactive equivalence proving. \\

& & Rtl2lean~\cite{rtl2lean26}
& Translates RTL into Lean models and constructs correctness theorems. \\

& & \textcolor{workblue}{CircuitProver~\cite{circuitprover26}}
& Automates hardware property proving through compositional reasoning and reusable Lean proofs. \\

\midrule

\multirow{4}{*}{\textbf{\shortstack[l]{Reasoning\\Engines}}}
& \multirow{3}{*}{SAT solving}
& AutoSAT; AutoModSAT~\cite{autosat24,automodsat26}
& Discovers and optimizes SAT solver heuristics through LLM-driven search. \\

& & AESAT~\cite{aesat26}
& Combines LLM-guided heuristic optimization with evolutionary search. \\

& & SATLUTION~\cite{satlution25}
& Extends algorithm discovery to repository-level SAT solver code evolution. \\

\cmidrule(lr){2-4}

& Model checking
& IC3-Evolve~\cite{ic3evolve26}
& Discovers reusable IC3/PDR heuristics through offline code evolution and correctness validation. \\

\bottomrule
\end{tabular}
\end{table*}

\textbf{AI-based Methods.}
Recent advances in LLMs create opportunities to support the construction, validation, and use of executable hardware specifications.
FormalRTL uses a supplied C reference model to guide RTL generation and correction through equivalence checking~\cite{formalrtl26}.
HINT introduces an executable hardware-intent representation with specification-level checking and correctness obligations for RTL lowering~\cite{hint26}.
These approaches demonstrate how executable specifications can provide explicit behavioral targets for AI-driven hardware development.

\textbf{Results and Insights.}
As illustrated in~\Cref{fig:spec_proof_comparison}(1), individual assertions may leave functional relationships unchecked even when all specified properties pass.
Executable reference models can describe broader input--state--output behavior and support equivalence or refinement checking.
Their effectiveness depends on accurately capturing designer intent, including state transitions, timing, and interface behavior.
Automating the construction and validation of reliable formal executable specifications remains an important research opportunity.

\subsection{Proof Construction: Agent-driven Formal Reasoning}
\label{sec:proof}

Agent-driven formal reasoning aims to improve the \textit{\textbf{automation, efficiency, and scalability}} of hardware formal verification.
AI agents use design semantics and verification feedback to guide proof search, construct supporting lemmas, and reuse verified knowledge.
We discuss two complementary directions: agentic model checking and agentic theorem proving.

\subsubsection{Agentic Model Checking}
\label{sec:agentic_mc}

Agentic model checking uses high-level hardware semantics and verification feedback to improve the efficiency of traditional model checking algorithms.

\textbf{Traditional Background.}
Model checking (MC) automatically reasons about hardware state spaces to verify correctness properties.
As illustrated in~\Cref{fig:formal_taxonomy}, verification can cover bounded or unbounded executions.
Bounded model checking (BMC) searches for counterexamples within a finite depth using SAT/SMT solving~\cite{biere1999symbolic,niemetz2018btor2}.
Unbounded techniques, including $k$-induction and IC3/PDR, establish correctness over all reachable states through inductive reasoning~\cite{sheeran2000checking,yu2022stratified,een2011efficient,goel2019model}.
These methods provide a high degree of automation, but their efficiency often depends on discovering useful invariants and managing large state spaces.

\textbf{AI-based Methods.}
Recent approaches introduce LLM-generated semantic guidance through two strategies.
(1) \emph{Pre-search guidance:}
LLM4PDR generates predicates, clauses, and helper assertions to guide word-level PDR, with candidate hints validated by the underlying model checker~\cite{llm4pdr26}.
(2) \emph{Feedback-driven guidance:}
CIll iteratively generates supporting invariants from counterexamples to induction (CTIs), while Agentic-IC3 interacts with evolving proof states to propose semantic lemmas and guide clause generalization~\cite{cill26,agenticic3_26}.
These approaches allow model checking to incorporate design-specific semantic knowledge during proof search.

\textbf{Results and Insights.}
Recent studies report improved verification performance on challenging arithmetic and processor designs using LLM-generated semantic guidance~\cite{llm4pdr26,cill26,agenticic3_26}.
High-level RTL semantics can help identify useful invariants and functional relationships that are difficult to discover through conventional search alone.
Pre-generated hints introduce limited inference overhead, while interactive agents can adapt to evolving proof states.
Balancing semantic guidance quality, proof-search efficiency, and inference cost remains an important challenge for larger designs.\looseness=-1

\subsubsection{Agentic Theorem Proving}
\label{sec:agentic_tp}

Agentic theorem proving automates machine-checked correctness proofs through proof planning, lemma generation, and compositional reasoning.
An emerging direction is to accumulate and reuse verified proof knowledge across hardware verification tasks.

\textbf{Traditional Background.}
Theorem proving (TP) constructs machine-checked correctness proofs using proof assistants such as Coq and Lean.
As illustrated in~\Cref{fig:formal_taxonomy}, theorem proving supports mathematical induction, expressive specifications, and compositional reasoning~\cite{kami17,hunt2017industrial}.
For example, Kami uses Coq to establish modular correctness properties for parameterized hardware designs~\cite{kami17}.
Compared with automated state-space exploration, theorem proving allows explicit reasoning over abstract state relations, parameterized designs, and reusable lemmas.
However, traditional theorem proving requires substantial human expertise in formal modeling, proof decomposition, and interactive proof construction.

\textbf{Agentic Methods.}
Recent approaches use LLM agents to automate hardware formalization and proof construction.
CktFormalizer combines typed circuit representations with interactive theorem proving, while Rtl2lean translates RTL into Lean models and constructs hierarchical correctness theorems~\cite{cktformalizer26,rtl2lean26}.
These approaches use proof-assistant feedback to guide lemma generation, proof decomposition, and iterative repair.
They also create opportunities for compositional proving and reusable proof libraries.

\textbf{Our Work.}
CircuitProver~\cite{circuitprover26} combines hardware-aware proof planning with a reusable Lean library for circuit property verification.
The agent generates supporting lemmas, repairs failed proofs using Lean feedback, and reuses verified theorems and proof strategies.
Compositional reasoning enables the construction of higher-level correctness proofs from verified component properties.\looseness=-1

\textbf{Results and Insights.}
CircuitProver~\cite{circuitprover26} completes all 63 evaluated tasks, compared with 58 for its baseline, while reducing average proof rounds from 9.2 to 4.6.
These results demonstrate the benefits of structured proof construction and reusable proof knowledge.
As hardware complexity increases, automatically discovering inductive invariants, intermediate lemmas, and compositional proof structures becomes increasingly important.
A promising direction is to develop agentic proving systems that continuously build and reuse machine-checked hardware proof libraries.

\subsection{Reasoning Engines: Agentic Algorithm Discovery}
\label{sec:engines}

Developing efficient reasoning engines traditionally requires substantial expertise in algorithm design, implementation, and optimization.
Recent advances in LLM coding agents enable automated algorithm discovery through iterative code modification, benchmarking, and correctness validation.
We discuss two directions: SAT solving and model checking, highlighting the trend from heuristic optimization toward autonomous algorithm development.

\subsubsection{SAT Solving}
\label{sec:sat_engine}

Agentic SAT solving is evolving from automated heuristic tuning toward broader algorithm discovery and repository-level code evolution.

\textbf{Traditional Background.}
SAT solving provides a fundamental reasoning capability for hardware formal verification, including bounded model checking, equivalence checking, and inductive proof procedures.
As illustrated in~\Cref{fig:formal_taxonomy}, SAT/SMT solvers discharge logical constraints generated by verification algorithms~\cite{niemetz2018btor2,niemetz2023bitwuzla}.
Modern SAT solvers commonly use conflict-driven clause learning (CDCL), supported by heuristics for variable branching, restarts, and clause management.
These heuristics strongly affect solving performance and traditionally require extensive expert knowledge and manual optimization.

\textbf{Agentic Methods.}
Recent approaches use LLMs to discover and optimize SAT solver heuristics through iterative code generation and evaluation.
AutoSAT explores heuristic generation within a modular CDCL solver, while AutoModSAT and AESAT combine LLM-guided optimization with evolutionary search~\cite{autosat24,automodsat26,aesat26}.
Beyond individual heuristic functions, SATLUTION uses planning and coding agents to evolve larger solver codebases, with distributed benchmarking and accumulated optimization knowledge guiding subsequent iterations~\cite{satlution25}.
These methods expand the search space from predefined parameters to algorithmic implementations and interactions among solver components.

\begin{figure*}[t]
\centering
\vspace{-.2in}
\includegraphics[width=0.9\linewidth]{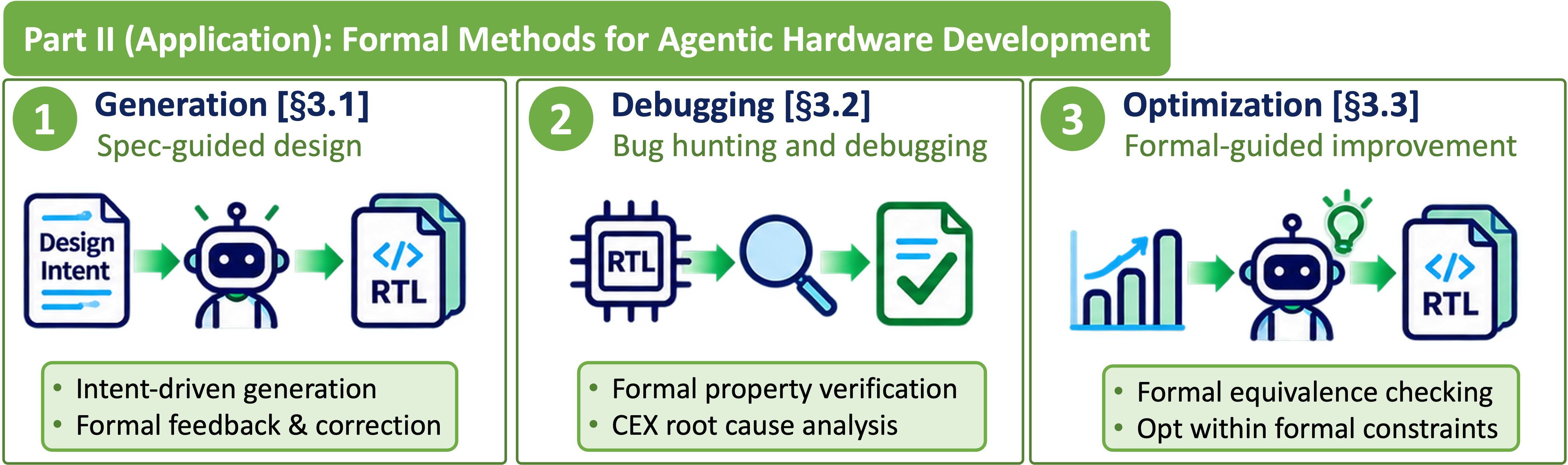}
\vspace{-.15in}
\caption{Three applications of formal methods in agentic hardware development.
Formal verification guides RTL generation through correctness feedback, supports debugging through counterexample-based diagnosis and repair, and enables correctness-preserving design optimization.}
\vspace{-.2in}
\label{fig:part2}
\end{figure*}

\textbf{Results and Insights.}
LLM-driven algorithm discovery has produced competitive SAT solvers, including AESAT's AE-Kissat-MAB, which won the 2025 SAT Competition main track~\cite{aesat26}.
AutoSAT, AutoModSAT, and SATLUTION also report improvements over their evaluated baselines~\cite{autosat24,automodsat26,satlution25}.
These results suggest that \emph{agentic code evolution can explore algorithmic choices beyond manually designed search spaces}.
Repository-level evolution further enables the joint optimization of interacting solver components.
An important research challenge is to discover improvements that generalize across diverse SAT instances while maintaining solver correctness.

\subsubsection{Model Checking}
\label{sec:mc_engine}

Agentic algorithm discovery is extending from SAT solving to model checking, where verification performance depends on more complex state-space reasoning and search strategies.

\textbf{Traditional Background.}
Model checking engines combine symbolic state-space algorithms with SAT/SMT solving to discharge verification obligations.
As illustrated in~\Cref{fig:formal_taxonomy}, their reasoning capabilities support automated correctness checking over bounded or unbounded executions.
Algorithms such as IC3/PDR rely on several interacting heuristic decisions, including inductive generalization, proof-obligation scheduling, and clause propagation~\cite{een2011efficient,goel2019model}.
These choices strongly influence verification efficiency and traditionally require substantial expertise in algorithm design and optimization.

\textbf{Agentic Methods.}
Recent research explores offline LLM-driven evolution of model checking algorithms.
IC3-Evolve~\cite{ic3evolve26} generates and evaluates code modifications to IC3/PDR heuristics through iterative experimentation.
Candidate implementations are evaluated on hardware benchmarks, with independently checked proof certificates and replayable counterexamples validating their results.
The discovered heuristics are integrated into a standalone model checker without runtime LLM inference.

\textbf{Results and Insights.}
IC3-Evolve~\cite{ic3evolve26} demonstrates the potential of automated algorithm discovery for hardware model checking.
Compared with task-specific proof guidance, offline algorithm discovery produces reusable improvements that can benefit future verification tasks.
Current exploration mainly focuses on selected heuristic components, leaving opportunities for broader algorithm discovery in state-space abstraction, invariant generalization, and interactions with SAT/SMT solvers.
\emph{Correctness validation, search efficiency, and cross-design generalization} are key challenges for developing reliable self-improving model checkers.

\section{Part II: Formal Methods for Agentic Hardware Development}
\label{sec:development}

Agentic hardware development involves three closely related activities: generating hardware implementations, debugging functional errors, and optimizing designs for better power, performance, and area (PPA).
Each activity requires formal verification in a different role.
As illustrated in~\Cref{fig:part2},
(1) \emph{Generation} (\Cref{sec:development_generation}) uses formal specifications and verification feedback to guide RTL construction and correction;
(2) \emph{Debugging} (\Cref{sec:development_debugging}) uses counterexamples and formal reasoning to identify root causes and validate repairs;
and (3) \emph{Optimization} (\Cref{sec:development_optimization}) uses equivalence checking to explore alternative implementations while preserving functional correctness.
Together, these activities form an iterative design process in which formal verification provides continuous correctness feedback and constraints for agentic decision-making.
Table~\ref{tab:development_works} summarizes representative research in these directions.

\subsection{Generation: Formal-guided RTL Construction}
\label{sec:development_generation}

Formal-guided RTL generation is evolving from one-shot code generation toward iterative construction and correction using verification feedback.
Assertion properties, executable reference models, and verified intermediate representations provide complementary ways to establish correctness targets.

\textbf{Traditional Background.}
Traditional RTL development relies on engineers to translate design requirements into implementations and identify functional errors through simulation and verification.
LLMs automate RTL construction, increasing the need for reliable correctness feedback and automated correction.

\textbf{Formal-guided Methods.}
Recent approaches integrate formal verification into RTL generation through two directions.
(1) \emph{Property-guided generation:}
Formal properties provide correctness feedback for training data selection, error detection, and RTL correction.
OpenLLM-RTL uses formal verification to select training examples~\cite{openllmrtl24}.
(2) \emph{Formal-model-guided generation:}
Executable reference models and formally defined intermediate representations provide behavioral constraints and proof obligations for hardware generation.
FormalRTL uses a supplied C reference model to guide RTL generation and correction through equivalence checking~\cite{formalrtl26}.
CktFormalizer uses typed circuit representations and interactive theorem proving to support hardware generation~\cite{cktformalizer26}.
HINT introduces an executable hardware-intent representation with pre-RTL checking and correctness obligations for RTL lowering~\cite{hint26}.
Proof2Silicon uses formally verified Dafny programs to guide hardware generation~\cite{proof2silicon25}.

\begin{figure}[t]
    \centering
    \includegraphics[width=0.99\linewidth]{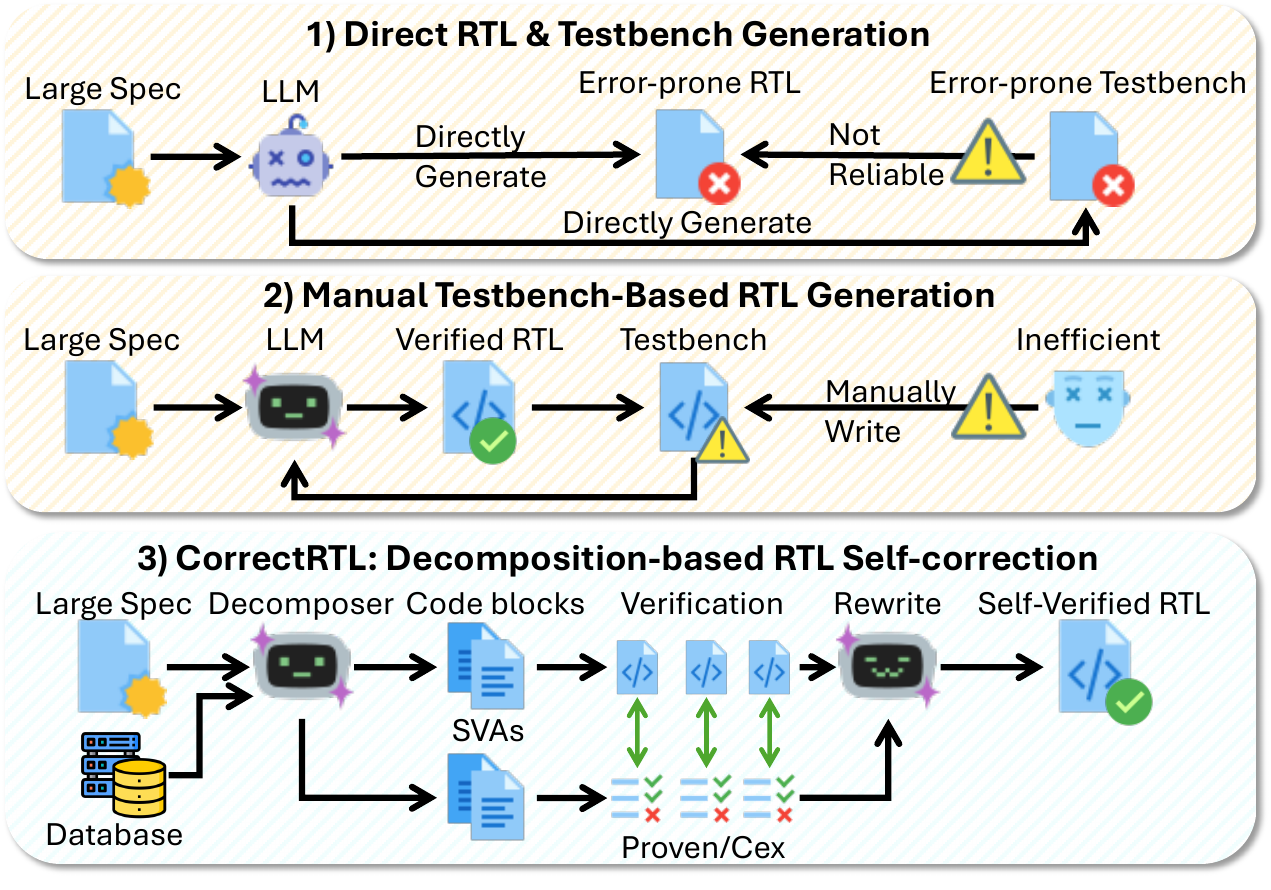}
    \vspace{-.1in}
    \caption{Overview of our CorrectRTL~\cite{correctrtl26} and its assertion-guided RTL self-correction workflow.}
    \label{fig:correct-rtl-figure}
    \vspace{-.2in}
\end{figure}

\textbf{Our Work.}
CorrectRTL~\cite{correctrtl26} introduces an SVA-based self-correction framework for LLM-generated RTL.
As illustrated in~\Cref{fig:correct-rtl-figure}, it uses automatically generated assertions to provide formal feedback for RTL correction.
The verification results identify violated requirements, and an LLM agent iteratively repairs the implementation.
This reduces dependence on manually constructed testbenches during the correction process.

\definecolor{workgreen}{RGB}{0,125,75}

\begin{table*}[t]
\centering
\vspace{-.2in}
\caption{Representative research in formal methods for agentic hardware development.}
\label{tab:development_works}
\vspace{-.1in}

\small
\setlength{\tabcolsep}{4pt}
\renewcommand{\arraystretch}{1.15}

\begin{tabular}{@{}
p{0.11\textwidth}
p{0.13\textwidth}
p{0.18\textwidth}
p{0.53\textwidth}
@{}}
\toprule
\textbf{Category} &
\textbf{Direction} &
\textbf{Representative Work} &
\textbf{Key Contribution} \\
\midrule

\multirow{6}{*}{\textbf{Generation}}
& \multirow{2}{*}{\shortstack[l]{Property-guided\\generation}}
& \textcolor{workgreen}{OpenLLM-RTL~\cite{openllmrtl24}}
& Uses formal verification to select training examples for RTL generation. \\

& & \textcolor{workgreen}{CorrectRTL~\cite{correctrtl26}}
& Uses generated assertions and formal feedback for iterative RTL correction. \\

\cmidrule(lr){2-4}

& \multirow{4}{*}{\shortstack[l]{Formal-model-\\guided generation}}
& FormalRTL~\cite{formalrtl26}
& Uses C reference models for equivalence-guided RTL generation and correction. \\

& & CktFormalizer~\cite{cktformalizer26}
& Uses typed circuit representations and theorem proving for hardware generation. \\

& & HINT~\cite{hint26}
& Introduces an executable hardware-intent representation with pre-RTL checking and RTL-lowering obligations. \\

& & Proof2Silicon~\cite{proof2silicon25}
& Uses verifier-guided learning and formally verified intermediate programs for hardware generation. \\

\midrule

\multirow{5}{*}{\textbf{Debugging}}
& \shortstack[l]{Failure\\diagnosis}
& FVDebug~\cite{fvdebug25}
& Uses formal counterexamples and causal reasoning for root-cause analysis. \\

\cmidrule(lr){2-4}

& \multirow{4}{*}{\shortstack[l]{Formal-guided\\repair}}
& AssertSolver~\cite{assertsolver25}
& Uses assertion failures and repair feedback to guide RTL correction. \\

& & Stitch~\cite{stitch26}
& Uses counterexample-guided RTL repair with formal validation. \\

& & Tran et al.~\cite{tran2026formalrepair}
& Combines open-source model checking with counterexample-guided RTL repair. \\

& & Das et al.~\cite{das2026multilinerepair}
& Uses assertion-guided repair and JasperGold to validate multi-line RTL fixes. \\

\midrule

\multirow{4}{*}{\textbf{Optimization}}
& \multirow{3}{*}{\shortstack[l]{RTL-level\\optimization}}
& ROVER~\cite{rover24}
& Explores equivalent logic transformations through verified e-graph rewriting. \\

& & ASPEN~\cite{aspen25}
& Combines LLM rewrite rules with equivalence checking and PPA feedback. \\

& & \textcolor{workgreen}{Dr.~RTL~\cite{dr_rtl26}}
& Uses timing-guided agentic RTL optimization with sequential equivalence checking. \\

\cmidrule(lr){2-4}

& \shortstack[l]{Layout-level\\optimization}
& \textcolor{workgreen}{E-Layout~\cite{elayout26}}
& Enables equivalence-preserving layout logic restructuring using physical feedback. \\

\bottomrule
\end{tabular}
\end{table*}

\textbf{Results and Insights.}
CorrectRTL improves Pass@1 from 20.5\% to 49.0\% on the 200-design ScaleBench benchmark, with correctness independently evaluated through sequential equivalence checking~\cite{correctrtl26}.
These results demonstrate how formal feedback can improve RTL generation through iterative correction.
Property-guided approaches provide targeted feedback on specified behaviors, while executable models and formal intermediate representations support more structured hardware generation.
The effectiveness of these methods depends on the reliability of their formal specifications and the correctness of the transformations connecting them to RTL implementations.

\subsection{Debugging: Formal-guided Diagnosis and Repair}
\label{sec:development_debugging}

Formal-guided debugging is progressing from interpreting verification failures toward automated root-cause diagnosis and formally validated repair.
Counterexamples and assertion failures provide concrete evidence for guiding agent decisions.

\textbf{Traditional Background.}
Hardware debugging traditionally requires engineers to analyze waveforms, simulation traces, and formal counterexamples to identify functional errors.
Root-cause localization and repair are particularly challenging for sequential designs, where failures may depend on interactions across multiple cycles.\looseness=-1

\textbf{Formal-guided Methods.}
Recent approaches explore two complementary capabilities.
(1) \emph{Failure diagnosis:}
AI agents analyze formal counterexamples, design dependencies, and specifications to identify likely root causes.
FVDebug combines counterexample analysis with causal reasoning and RTL context to explain formal verification failures and suggest fixes~\cite{fvdebug25}.
(2) \emph{Formal-guided repair:}
Agents use failing properties and counterexamples to propose RTL modifications, followed by formal rechecking.
AssertSolver and Stitch explore assertion- and counterexample-guided correction~\cite{assertsolver25,stitch26}.
An open-source multi-agent workflow~\cite{tran2026formalrepair} further combines SymbiYosys and Z3 with counterexample-driven RTL repair.
Recent work~\cite{das2026multilinerepair} on RISC-V processors uses assertion-guided LLM repair and JasperGold verification to address multi-line RTL bugs.\looseness=-1

\textbf{Results and Insights.}
These studies show how formal verification can support both fault localization and automated correction.
Counterexamples reveal failing behavior, while repeated model checking evaluates whether proposed repairs satisfy the specified properties.
Recent work also extends formal-guided repair toward more complex, multi-line bugs.
The main challenges include accurate root-cause diagnosis, efficient repair search, and sufficient specification coverage to prevent incomplete or misleading fixes.

\subsection{Optimization: Correctness-preserving Design Exploration}
\label{sec:development_optimization}

\begin{figure}[t]
\centering
\includegraphics[width=0.8\linewidth]{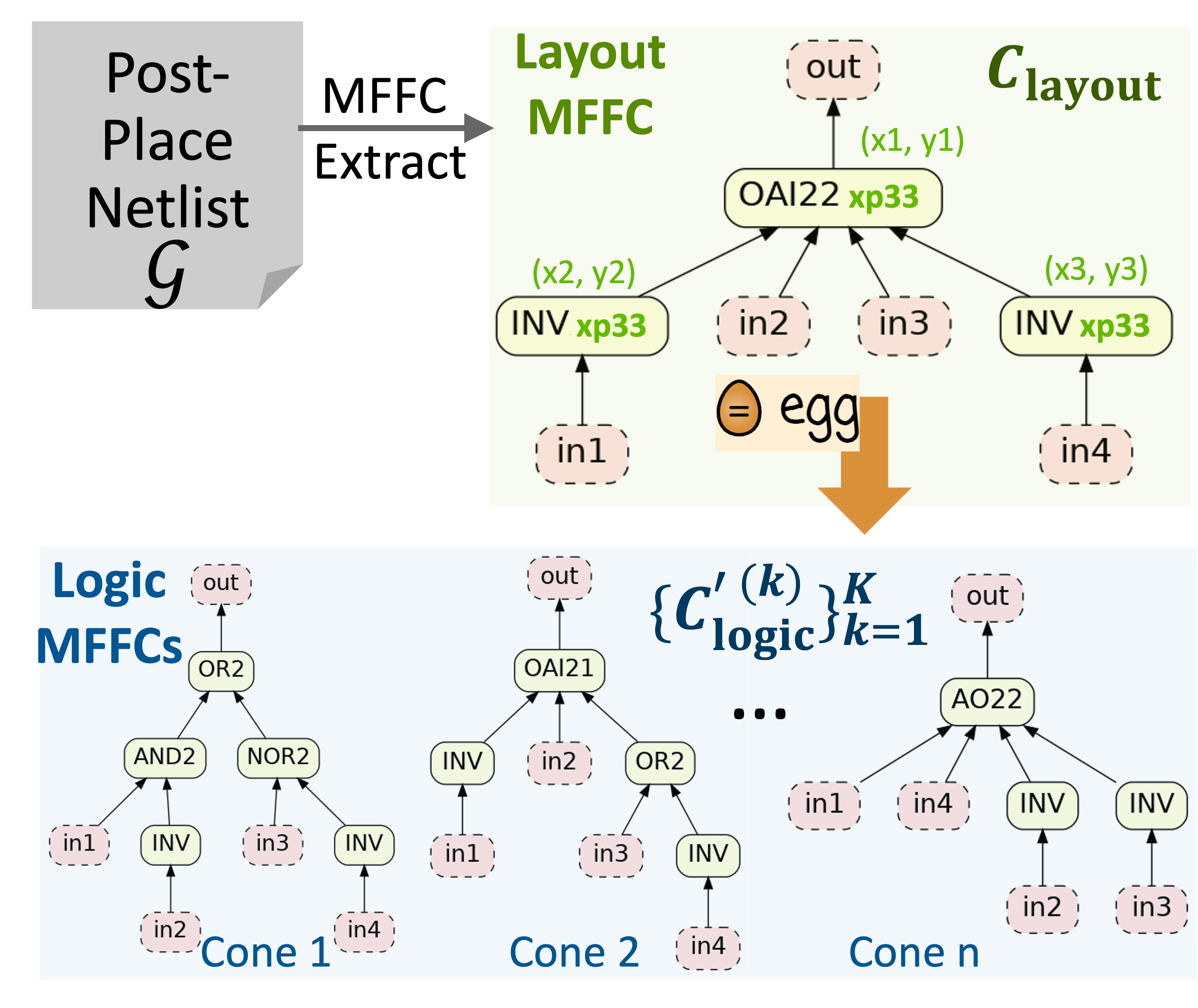}
\vspace{-.2in}
\caption{Our E-Layout~\cite{elayout26} uses e-graph equality saturation to explore formal equivalent logic structures for timing optimization.}
\vspace{-.2in}
\label{fig:elayout}
\end{figure}

Formal equivalence checking enables AI agents to explore alternative implementations under explicit correctness constraints.
Recent approaches combine formal checking with EDA feedback to optimize hardware at both RTL and physical-design stages.

\textbf{Traditional Background.}
Traditional synthesis and RTL optimization use Boolean rewriting, restructuring, and technology mapping to improve power, performance, and area (PPA).
Formal equivalence checking validates correctness-preserving transformations.
At later design stages, placement and timing information provide additional guidance for selecting effective logic transformations.\looseness=-1

\textbf{Formal-guided Methods.}
Recent approaches explore two directions.
(1) \emph{Verified logic rewriting:}
ROVER uses verified e-graph rewriting, while ASPEN combines LLM-generated rewrite rules with equivalence checking and PPA feedback~\cite{rover24,aspen25}.
(2) \emph{Agentic design exploration:}
AI agents iteratively generate design alternatives, evaluate their implementation quality using EDA tools, and validate functional equivalence.
This supports design-specific optimization across a broader range of transformations.

\textbf{Our Work.}
Our research explores correctness-preserving optimization at two design stages.
(1) \emph{RTL-level optimization:}
Dr.~RTL~\cite{dr_rtl26} uses timing reports and EDA feedback to guide agentic RTL transformations, with sequential equivalence checking (SEC) validating functional correctness.
(2) \emph{Layout-level optimization:}
E-Layout~\cite{elayout26} explores post-placement logic restructuring through equality saturation and layout-aware structure selection.
As illustrated in~\Cref{fig:elayout}, e-graph rewriting generates multiple equivalent implementations of a logic cone, expanding the structural optimization space.
Learning-based selection uses placement and timing information to identify promising alternatives.
Local SMT checking and full-design equivalence verification validate the transformations.

\textbf{Results and Insights.}
Across 20 designs, Dr.~RTL achieves average improvements of 21.3\% in worst negative slack (WNS) and 16.9\% in total negative slack (TNS), together with a 5.8\% area reduction and an 86\% SEC pass rate~\cite{dr_rtl26}.
E-Layout reports average improvements of 6.07\% in TNS and 2.29\% in WNS, together with a 1.07\% area reduction across nine designs~\cite{elayout26}.
These results highlight how formal equivalence enables broader structural exploration, while physical feedback guides the selection of implementations with better PPA.\looseness=-1

\section{Future Research Directions}
\label{sec:discussion}
\begin{figure}[t]
\centering
\includegraphics[width=1\linewidth]{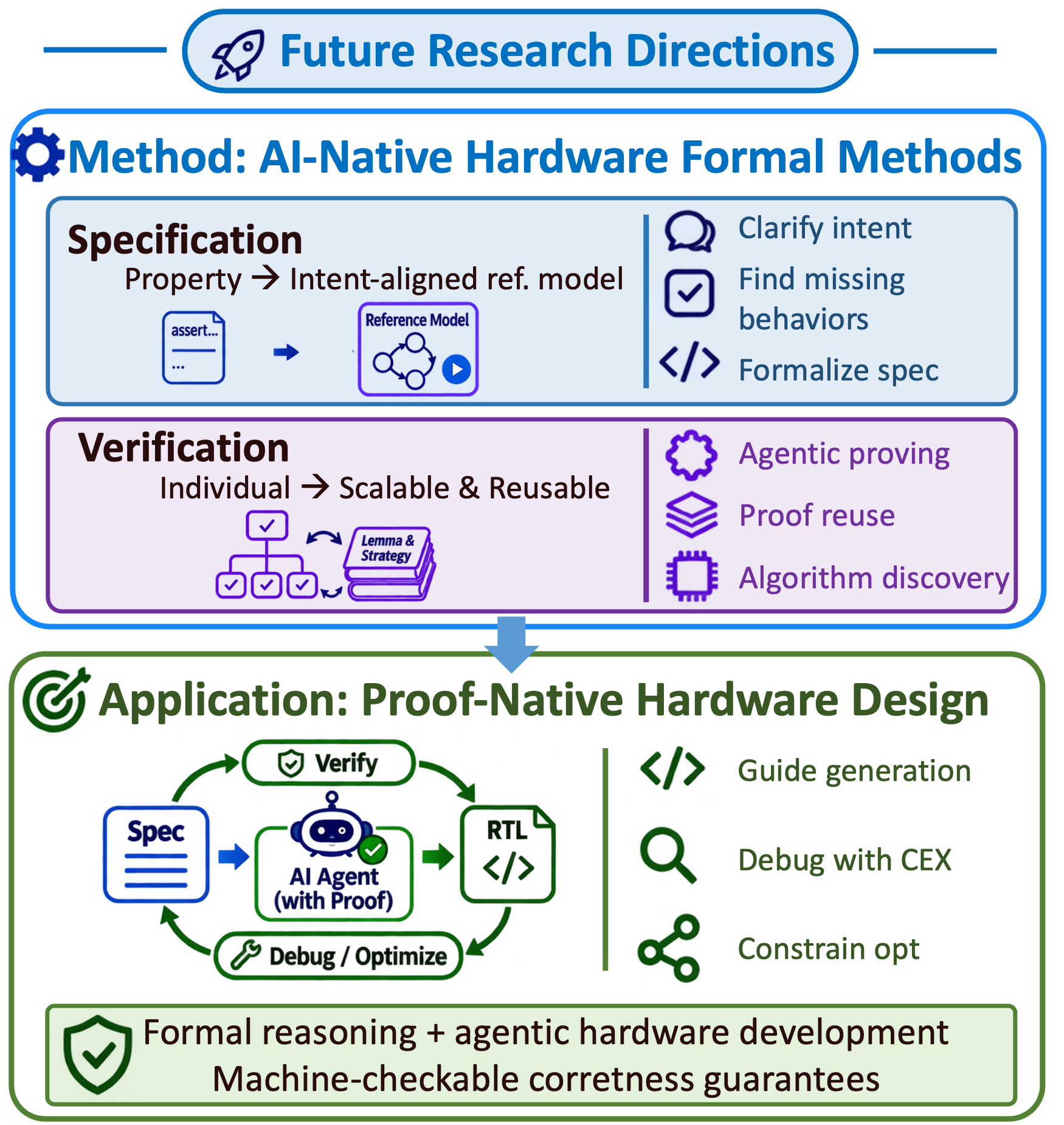}
\vspace{-.2in}
\caption{Future research directions toward scalable AI-native formal methods and proof-native hardware development.}
\label{fig:future}
\end{figure}

As AI becomes more capable, formal methods are becoming both more important and more accessible.
As illustrated in~\Cref{fig:future}, we envision three research directions spanning formal verification methodology and its applications toward proof-native hardware development.

\subsection{Specification: From Isolated Properties to Intent-Aligned Reference Models}

As AI increasingly automates RTL generation, accurately capturing designer intent becomes more important.
A promising direction is to move from individual assertion properties toward more complete, executable formal reference models that support equivalence and refinement checking.
This shift highlights the growing importance of requirements engineering in hardware development.
AI agents can help designers clarify ambiguous requirements, identify missing behaviors, and translate design intent into formal specifications.
Ensuring that these specifications accurately reflect designer intent will be essential for reliable AI-native hardware development.

\subsection{Verification: From Individual Proofs to Scalable and Reusable Proving}

AI-native verification techniques offer opportunities to improve proof automation, efficiency, and scalability.
Agentic theorem proving can automate complex hardware proofs through invariant generation, proof decomposition, and interactive reasoning.
Compositional verification and proof reuse can further improve scalability by allowing agents to accumulate and reuse machine-checked lemmas and proof strategies across related designs.
Agentic algorithm discovery can also improve SAT/SMT solvers and model checking engines, making formal reasoning more efficient.

\subsection{Application: From Formal Verification Feedback to Proof-Native Design}

Formal verification is becoming an important part of AI-driven hardware generation, debugging, and optimization.
The next step is to integrate formal reasoning throughout the design process, maintaining correctness guarantees as AI agents continuously modify implementations.
Formal specifications can guide generation, counterexamples can support debugging, and equivalence checking can enable broader design exploration.
\textbf{Proof-native hardware development} connects these capabilities into a unified workflow, where AI agents generate, verify, and optimize hardware while maintaining machine-checkable correctness guarantees.

\section{Conclusion}
\label{sec:conclusion}

This paper presented a unified view of the growing relationship between AI and formal methods in the era of agentic hardware development.
We reviewed how AI advances formal specification, proof construction, and reasoning engines, and how formal methods support reliable RTL generation, debugging, and optimization.
Through representative research and our recent contributions, we highlighted the opportunities for improving verification automation, efficiency, and scalability.
As AI becomes more capable, the demand for correctness guarantees continues to grow, while the barriers to developing and applying formal methods are decreasing.
We envision formal methods becoming a native part of AI-driven hardware development, enabling increasingly autonomous design with machine-checkable correctness guarantees.

\section{ACKNOWLEDGEMENT}
This work is supported by Hong Kong Research Grants Council (RGC) CRF-YCRG C6003-24Y, 16216825, T46-415/25-R, and JLFS-YSF/E-605/26. It was partially conducted by ACCESS – AI Chip Center for Emerging Smart Systems, supported by the InnoHK initiative of the Innovation and Technology Commission of the Hong Kong Special Administrative Region Government.
\clearpage
\bibliographystyle{ACM-Reference-Format}
\bibliography{section/survey_references}
\end{document}